\pdfoutput=1
\documentclass[11pt, logo, a4paper, copyright, nonumbering]{nvidiatechreport}

\usepackage{hyperref}
\usepackage{url}
\usepackage{colortbl}
\usepackage{nicefrac}       %
\usepackage{microtype}      %
\usepackage[dvipsnames]{xcolor}         %
\usepackage{multirow}
\usepackage{multicol}
\usepackage{graphicx}
\usepackage{xspace}
\usepackage{amsmath}
\usepackage{adjustbox}
\usepackage{tcolorbox}
\usepackage{amssymb}
\usepackage{enumitem}
\usepackage{wrapfig}
\usepackage{xcolor}
\usepackage{subcaption}    %
\usepackage{float}
\usepackage{stfloats}
\usepackage{amsmath}
\usepackage{listings}
\usepackage{mdframed}
\usepackage{tcolorbox}
\usepackage{arydshln}
\usepackage{booktabs} 
\usepackage{color,soul}
\usepackage{makecell}
\usepackage[authoryear]{natbib}
\usepackage{caption}
\usepackage{makecell}

\newcommand{\method}[0]{\mbox{\textsf{NitroGen}}\xspace}

\hypersetup{
  colorlinks=false,
  pdfborder={0 0 1}
}

\tcbuselibrary{listings,breakable}

\definecolor{pastelblue}{RGB}{173,216,230}
\definecolor{pastelyellow}{RGB}{255,253,208}
\definecolor{pastelpink}{RGB}{255,209,220}
\definecolor{pastelgreen}{RGB}{176,226,172}
\definecolor{pastellavender}{RGB}{230,230,250}

\definecolor{NvidiaGreen}{RGB}{118, 185, 0}
\hypersetup{citecolor=NvidiaGreen}
\sethlcolor{red!15}

\title{\method: An Open Foundation Model for Generalist Gaming Agents}

\author{
Loïc Magne$^{1\;*}$, Anas Awadalla$^{1\;2\;*}$, Guanzhi Wang$^{1\;3\;*\;\dagger}$ \\
\;Yinzhen Xu$^{1}$, Joshua Belofsky$^{4}$, Fengyuan Hu$^{1}$, Joohwan Kim$^{1}$\\
\;Ludwig Schmidt$^{2}$, Georgia Gkioxari$^{3}$, Jan Kautz$^{1}$ \\
\;Yisong Yue$^{3\:\dagger}$, Yejin Choi$^{1\;2\;\dagger}$, Yuke Zhu$^{1\;5\;\dagger}$, Linxi ``Jim'' Fan$^{1\;\dagger}$\\
\small{$^1$ NVIDIA, $^2$ Stanford, $^3$ Caltech, $^4$ UChicago, $^5$ UT Austin} \\
\small{${^*}$ Co-lead, ${^\dagger}$ Co-advise} \\
\par\vspace{0.25em}
\small\url{https://nitrogen.minedojo.org}
}

\begin{abstract}
\textbf{Abstract:}

We introduce \method, a vision-action foundation model for generalist gaming agents that is trained on 40,000 hours of gameplay videos across more than 1,000 games. 
We incorporate three key ingredients: 1) an internet-scale video-action dataset constructed by automatically extracting player actions from publicly available gameplay videos, 2) a multi-game benchmark environment that can measure cross-game generalization, and 3) a unified vision-action model trained with large-scale behavior cloning. 
\method exhibits strong competence across diverse domains, including combat encounters in 3D action games, high-precision control in 2D platformers, and exploration in procedurally generated worlds. It transfers effectively to unseen games, achieving up to 52\% relative improvement in task success rates over models trained from scratch.
We release the dataset, evaluation suite, and model weights to advance research on generalist embodied agents.
\end{abstract}

\begin{document}

\maketitle

\begin{abstract}
We introduce \method, a video-action foundation model for generalist gaming agents that is trained on 40,000 hours of gameplay videos across more than 1,000 games. 
We incorporate three key ingredients: 1) an internet-scale video-action dataset constructed by automatically extracting player actions from publicly available gameplay videos, 2) a multi-game benchmark environment that can measure cross-game generalization, and 3) a unified vision-action model trained with large-scale behavior cloning. 
\method exhibits strong competence across diverse domains, including combat encounters in 3D action games, high-precision control in 2D platformers, and exploration in procedurally generated worlds. It transfers effectively to unseen games, achieving up to 52\% relative improvement in task success rates over models trained from scratch.
We release the dataset, evaluation suite, and model weights to advance research on generalist embodied agents.
\end{abstract}
    
\section{Introduction}

\begin{figure}[t]
  \centering
  \includegraphics[width=\textwidth]{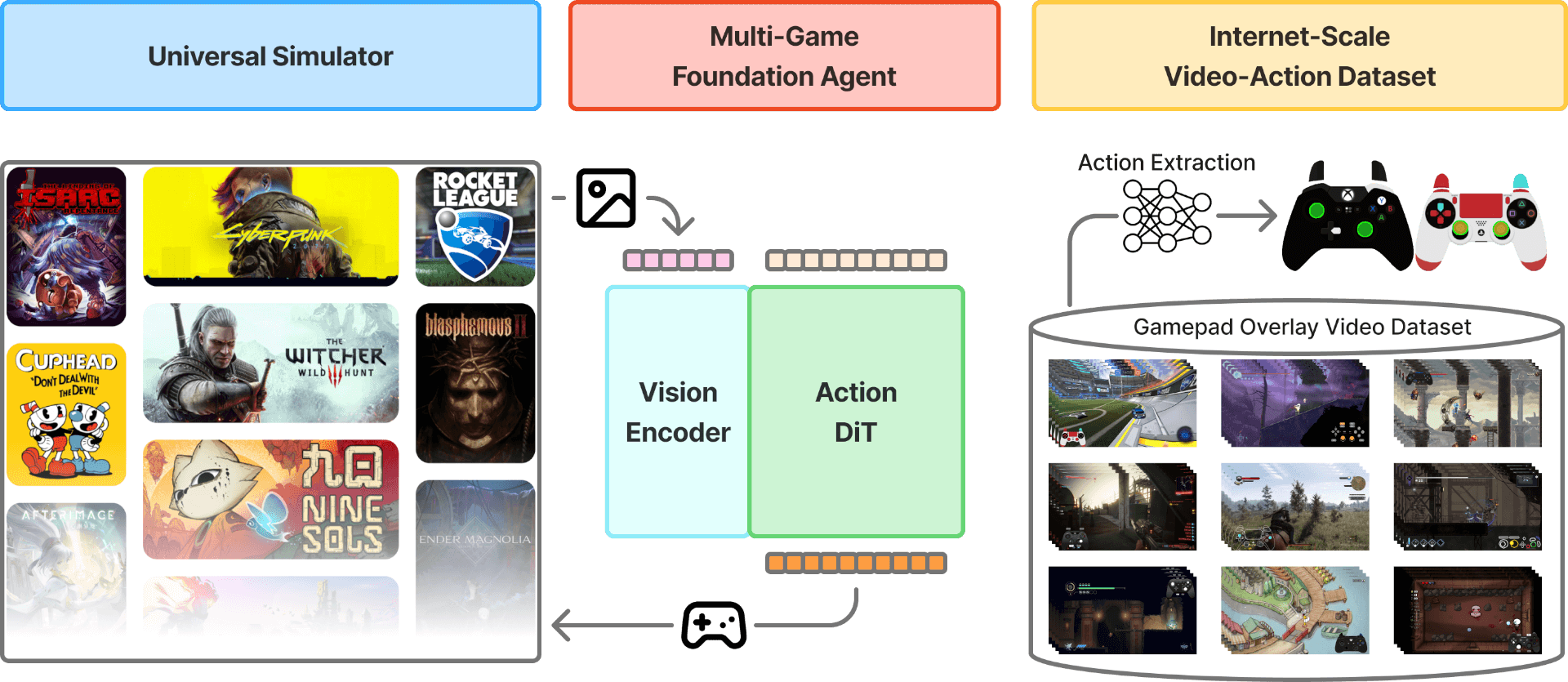}
  \caption{\textbf{\method overview.} \method consists of three main components: (1) \textbf{Multi-game foundation agent (center)} - a generalist vision-action model that takes in game observations and generates gamepad actions, enabling zero-shot gameplay across multiple titles and serving as a foundation for fine-tuning on new games; (2) \textbf{Universal simulator (left)} - an environment wrapper that allows any commercial game to be controlled through a Gymnasium API; and (3) \textbf{Internet-scale dataset (right)} - the largest and most diverse open-source gaming dataset curated from 40,000 hours of publicly available gaming videos, spanning more than 1,000 games with extracted action labels.}
  \label{fig:pull}
\end{figure}

Building generally capable embodied agents that can operate in unknown environments has long been considered a holy grail of AI research. While computer vision and large language models (LLMs) have achieved this generalization through large-scale pre-training on internet data \citep{brown2020languagemodelsfewshotlearners, devlin2019bertpretrainingdeepbidirectional, radford2021learningtransferablevisualmodels, dosovitskiy2021imageworth16x16words}, comparable progress in embodied AI has been impeded by the lack of large, diverse, and labeled action datasets.
Video games present an ideal domain for advancing embodied AI since they offer visually rich interactive environments and tasks that span a wide range of complexities and temporal horizons. However, prior approaches face substantial limitations.
\textbf{LLM–based methods} exploit either (1) hand-crafted programmatic APIs to expose internal game states and control agents \citep{wang2023voyager, volum-etal-2022-craft, wang2024describeexplainplanselect} or (2) complicated perception modules for textual information extraction and object detection \citep{tan2024cradle}. They enable complex task-solving but require complicated domain-specific design and tuning.
\textbf{Reinforcement learning} has achieved superhuman performance in individual games such as StarCraft II and Dota 2, but these agents are narrow, costly to train, and depend on specialized simulators rarely available for arbitrary games \citep{berner2019dota, silver2016mastering, vinyals2019grandmaster, mnih2013playing, mnih2015human}. \textbf{Behavior-cloning approaches} based on pixel observations have relied on expensive-to-collect demonstrations, constraining training to only a few game titles due to prohibitive data collection costs \citep{baker2022video,sima2024}.
To date, there has been little progress on developing open-source frameworks that can support the training and evaluation of generalist gaming agents, further hindering progress in this direction.

To address these limitations, we introduce \method, an open foundation model for video game environments trained on 40,000 hours of publicly available internet videos covering more than 1,000 games. We make three major contributions (Figure~\ref{fig:pull}):

\textbf{1. Internet-scale dataset of action-labeled videos.} 
We propose to use a new source of data from publicly available videos where content creators overlay their input commands in real time. We train an annotation model to extract frame-level actions with high accuracy, removing the need for costly manual data collection and capturing a wide spectrum of real player behaviors. Using this approach, we curate a dataset of 40,000 hours of video spanning more than 1,000 games, providing diverse demonstrations for large-scale training.

\textbf{2. Multi-task multi-game evaluation suite.}
To assess generalization in realistic settings, we design benchmark environments that comprise 30 tasks of varied complexity from 10 commercial games, covering diverse challenges such as combat, navigation, decision-making, platforming, exploration, and puzzle-solving. This benchmark reflects the demands of modern game environments, where agents must learn to adapt across heterogeneous mechanics and objectives. We provide a universal Gymnasium API \citep{towers2024gymnasium} for our evaluation suite that allows users to wrap any game to test diverse agent capabilities. This API is what we refer to as the \textbf{universal simulator} in Figure~\ref{fig:pull}.

\textbf{3. Large-scale behavior-cloning pre-training.} 
To demonstrate the feasibility and benefits of internet-scale pre-training, we train a vision-action transformer model on our dataset. We demonstrate strong results on our benchmark suite, validating our end-to-end pipeline and showing that it is possible to train a strong multi-game policy using only noisy internet data. We show the benefits of behavior-cloning pre-training by post-training our base model on games not seen during training. The model fine-tuned from the pre-trained \method weights shows up to $52\%$ relative improvement in success rates over the model trained from scratch, given a fixed data and compute budget.

We open-source the \method dataset, simulator, and pre-trained weights. We envision \method as a foundational resource that will enable the research community to accelerate progress toward building more generalist embodied agents, fostering new algorithms, model architectures, and applications in this emerging area.
  
\begin{figure}[t]
    \centering
    \begin{subfigure}[t]{\textwidth}
        \centering
        \includegraphics[width=\textwidth]{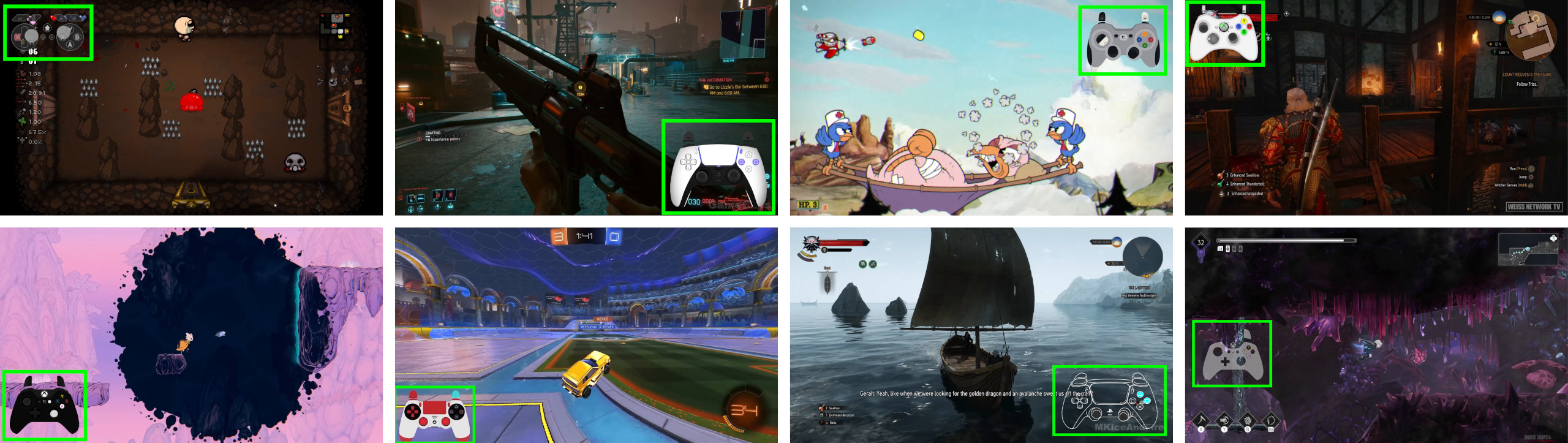}
        \caption{Examples of gamepad overlay videos.}
        \label{fig:gamepad_overlay}
    \end{subfigure}

    \vspace{1em}
    \begin{subfigure}[t]{\textwidth}
        \centering
        \includegraphics[width=\textwidth]{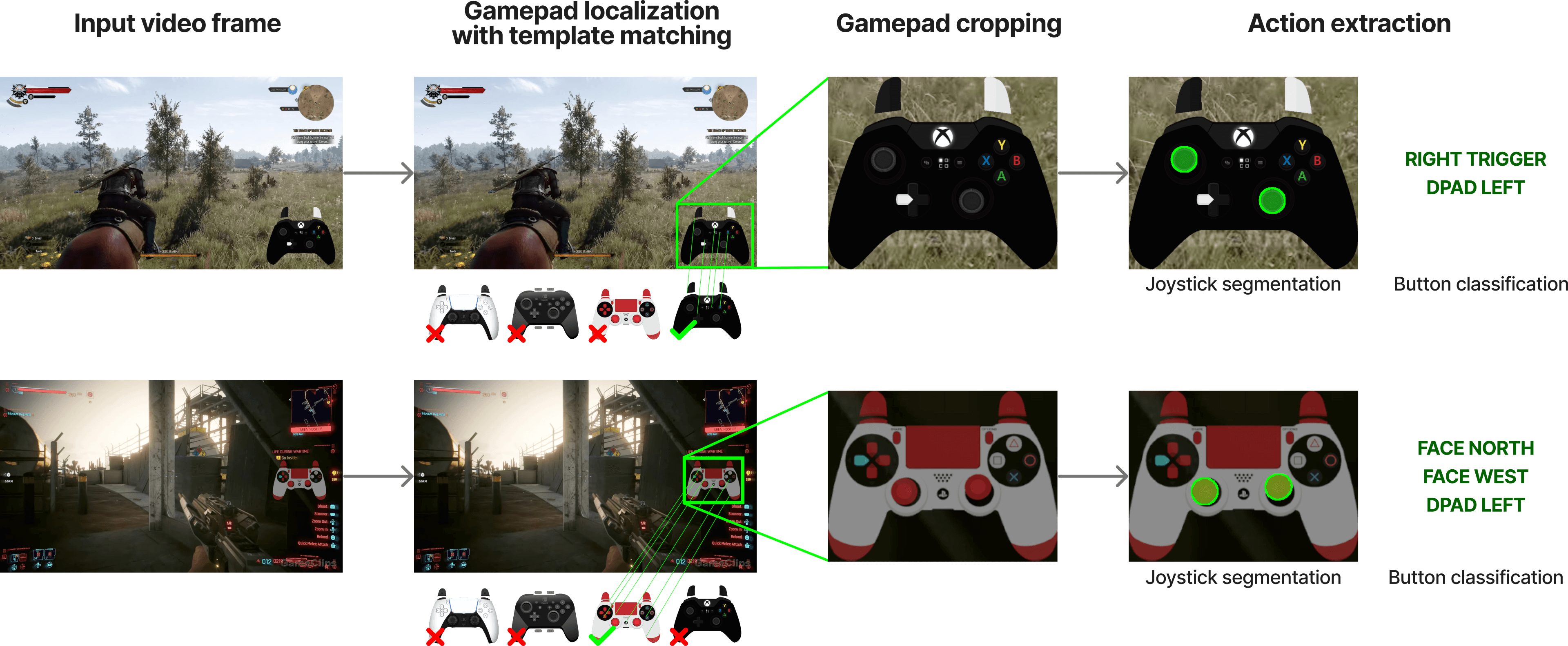}
        \caption{Action extraction pipeline.}
        \label{fig:action_extraction}
    \end{subfigure}
    
    \caption{\textbf{Video-action dataset pipeline overview.} We extract actions from on-screen displays which show the gamepad actions of the player in real-time; called ``input overlays''. 
    \textbf{(a) Dataset curation.} We collect publicly available videos displaying a “gamepad overlay”. The diversity of these overlays presents significant challenges, as gamepads vary widely across content creators in controller types (e.g., Xbox, PlayStation, or others), transparency levels, and visual artifacts introduced by video compression.
    \textbf{(b) Action extraction.} For each collected video, we localize the gamepad by sampling 25 frames and running \textbf{keypoint matching} against a curated set of templates using SIFT and XFeat features. We use the template-matching results to localize and crop the gamepad region from each video. A \textbf{hybrid classification–segmentation network} is then trained to predict joystick positions and button states from the cropped controller images, enabling accurate reconstruction of player inputs.}
    \label{fig:dataset}
\end{figure}

\section{Approach}

\method consists of three novel components: (1) an internet-scale video dataset with action labels, (2) a multi-game benchmark with a Gymnasium environment wrapper, and (3) a vision–action model pre-trained through large-scale behavior cloning. In this section, we provide details of each component.

\subsection{Internet-scale multi-game video-action dataset}
\label{sec:dataset}

\textbf{Annotation challenge.} A central challenge in training policies from internet videos is recovering the corresponding actions, since most gameplay recordings typically do not include the player’s inputs. We address this limitation by using a novel source of publicly available videos in which such labels can be recovered. These videos feature \textit{input overlay} software that displays a real-time visualization of the player’s actions, typically as a 2D image of a gamepad in a corner of the screen with pressed buttons highlighted (Figure \ref{fig:gamepad_overlay}).

\begin{figure}[t!]
    \centering
    \includegraphics[width=\textwidth]{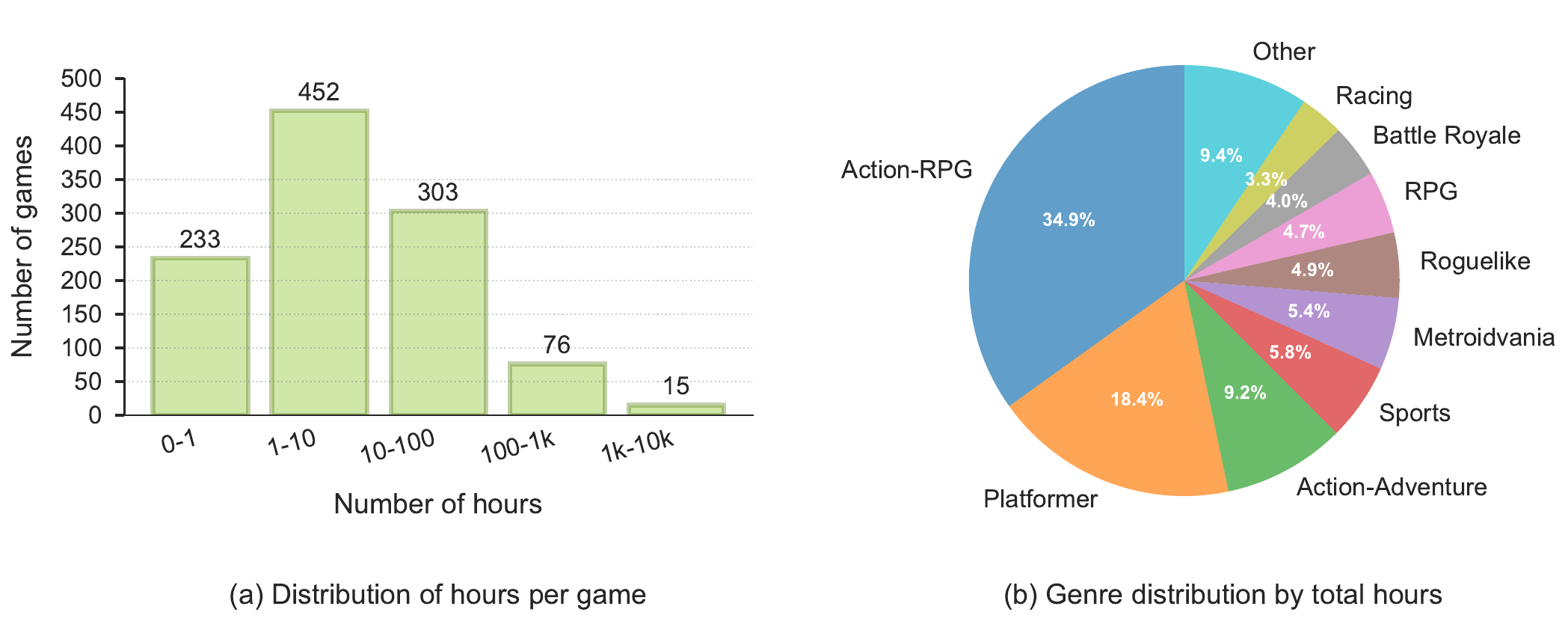}
    \caption{\textbf{Distribution of the \method dataset across games and genres.} After filtering, the \method dataset contains 40,000 hours of gameplay videos spanning more than 1,000 games. (a) Hours per game shows broad coverage, with 846 games having over one hour of data, 91 games with over 100 hours, and 15 games exceeding 1,000 hours each. (b) Genre distribution reveals Action-RPG games are most common (34.9\% of total hours), followed by Platformer (18.4\%) and Action-Adventure (9.2\%) games, with the remainder distributed across seven genres.}
    \label{fig:dataset_analysis}
\end{figure}

\textbf{Dataset curation.} 
Although input overlays appear in only a fraction of online gameplay videos, they occur frequently enough to enable the construction of a large-scale dataset. We collect 71,000 hours of raw video containing gamepad overlays. While input overlay software was originally used primarily within the speedrunning community, its use has since expanded to many action games and among both expert and casual players. To avoid over-representation of any single title, we use a combination of keyword-based searches and curation guided by content diversity, ensuring coverage across games, genres, and skill levels. This approach balances casual and competitive play styles while maintaining broad genre representation. Figure~\ref{fig:dataset_analysis} shows the distribution of gameplay hours by title and genre. 
The dataset covers more than 1,000 unique games, making it the largest labeled video–action dataset for video games to date. It contains 38,739 videos from 818 different content creators, with an average video duration of 1 hour and 50 minutes.

\textbf{Action extraction.}
We extract player inputs from gameplay videos through a three-stage pipeline: (1) template matching to locate and crop the gamepad overlay, (2) gamepad action parsing using a fine-tuned segmentation model, and (3) quality filtering to ensure accurate and meaningful data.

\textit{Stage 1: Template matching.} 
To locate gamepad overlays within gameplay videos, we apply template matching using a curated set of approximately 300 common controller templates. For each video, we sample 25 frames and perform feature matching with SIFT~\citep{sift_lowe2004distinctive} and XFeat~\citep{potje2024xfeat} against all curated templates. We estimate an affine transformation from the paired keypoints and require at least 20 inliers for a match to be considered valid. We then extract the region with the highest matching score, which defines the gamepad location for subsequent processing. Figure \ref{fig:action_extraction} shows examples of successful match.

\textit{Stage 2: Gamepad action parsing.}
We parse controller states using a fine-tuned SegFormer~\citep{Xie2021SegFormerSA} segmentation model that processes pairs of consecutive frames. The model takes two consecutive frames as input (concatenated along the spatial dimension) to capture short-term temporal dynamics. It outputs a segmentation mask to localize joystick positions on a discrete $11\times11$ grid, and binary button states (Figure~\ref{fig:action_extraction}). Empirically, we find that estimating joystick positions via segmentation masks significantly outperforms direct regression of joystick coordinates.

We train the annotation model using synthetic data generated by sampling frames from the \method training set and programmatically overlaying controller templates using the Open Joystick Display\footnote{\url{https://github.com/AkikoKumagara/open-joystick-display}}, Input Overlay\footnote{\url{https://github.com/univrsal/input-overlay}}, and GamePad Viewer\footnote{\url{https://beta.gamepadviewer.com/}} software. For each template, we produce multiple frames with random button states and joystick positions, yielding 8M labeled frames. To simulate real-world visual artifacts, we vary overlay opacity, controller size, and video compression, generating multiple variants per frame. We train the action parsing SegFormer model using the AdamW optimizer~\citep{adam_optimizer, loshchilov2017decoupled} with a learning rate of $0.0001$, linear learning rate decay, weight decay of $0.1$, and a batch size of $256$.

At inference, we compute precise joystick positions by detecting contours for each joystick over the entire video. To estimate the center position of each joystick, we average positions from all frames where the joystick is classified as centered in the $11\times11$ discrete grid. We then normalize the positions to the range $[-1.0, 1.0]$ using the 99th percentile of absolute $x$ and $y$ values over the video to reduce the influence of outliers.


\textit{Stage 3: Quality filtering.} 
The final stage applies targeted filtering strategies to ensure high-quality data. During training, we observe that using the raw 71,000 hours of data leads to the model over-predicting the null action as noted in VPT \cite{baker2022video}. To avoid that, we discard segments based on action density: we only keep chunks where at least 50\% of the timesteps have non-zero button or joystick actions, resulting in 55\% of the data being kept. For all gameplay videos, we mask the on-screen controller to prevent models from exploiting it as a shortcut. 

\begin{figure}[t]
  \centering
  \includegraphics[width=\textwidth]{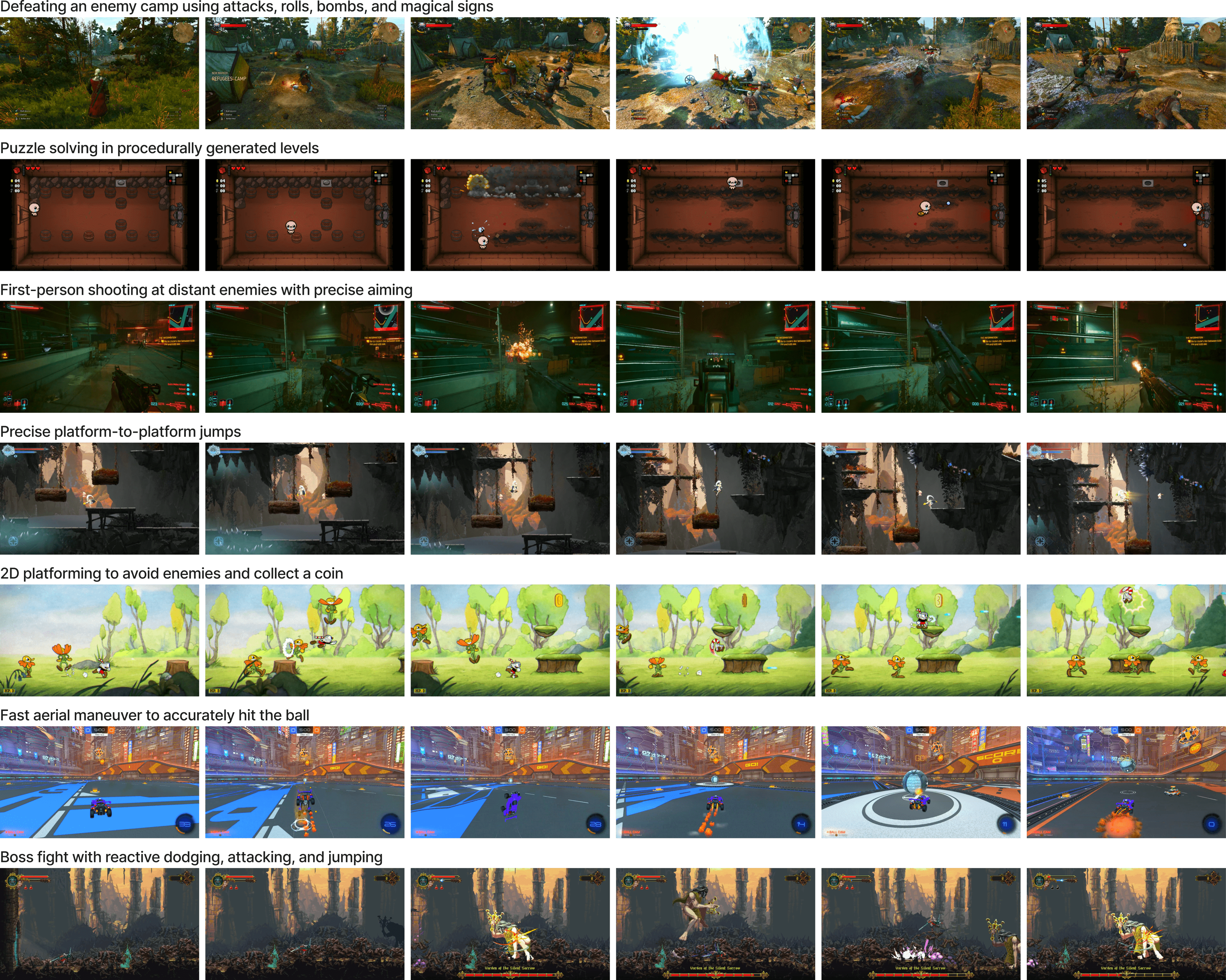}
  \caption{\textbf{In-game rollouts.} We show \method performing tasks in diverse 2D and 3D environments. These tasks can take from a few seconds to a few minutes to perform. Some of them include memorization, while others are performed in procedurally generated worlds and require the model to adapt.}
  \label{fig:rollout}
\end{figure}

\subsection{Evaluation suite}

\label{sec:gym_api}

\textbf{Universal simulator for any game title.}
Many research environments provide a Gymnasium API \citep{towers2024gymnasium} that enables programmatic control of the simulation. To bring this capability to commercial video games, which typically lack such an interface, we develop a universal simulator that can wrap any game title with a Gymnasium API for model development. The library intercepts the game engine’s system clock to control simulation time, enabling frame-by-frame interaction without modifying game code. This approach works with any title that uses the system clock for physics and interactions, which is a common practice in game development. We leave real-time or asynchronous deployment to future work. Frequent pausing and resuming during gameplay could potentially affect the game's physics engine in unknown ways, we verify that this process does not alter games' physics and behaviors (see Appendix~\ref{app:inference}.).


\textbf{Unified observation and action space.}
Using this simulator, we introduce a multi-game, multi-task benchmark with a shared interface across all titles. Observations are single RGB frames. Actions consist of a standardized 16-dimensional binary vector for gamepad buttons (4 d-pad buttons, 4 face buttons, 2 shoulders, 2 triggers, 2 joystick thumb buttons, start, back) plus a 4-dimensional continuous vector for joystick positions. Unlike prior work that defines game or task-specific action spaces~\citep{baker2022video,guss2019minerl}, this unified layout facilitates direct policy transfer across diverse games.


\textbf{Diverse evaluation tasks.}
The evaluation suite serves as a universal evaluation framework for multi-game policies, covering 10 games across diverse visual styles and genres with 30 tasks total. The suite includes five 2D games and five 3D games, each testing different skill combinations. The 2D games include three side-scrollers and two top-down roguelikes with procedurally generated levels. The 3D games consist of two open-world games, two combat-focused action-RPGs, and one sports game.

Tasks are distributed across three categories: 11 combat tasks (boss fights, enemy encounters), 10 navigation tasks (reaching specific locations, traversing environments), and 9 game-specific tasks (unique mechanics particular to individual games). Each task has clearly defined start and goal states, with attempts typically lasting a few minutes, though human players may require several hours of repeated attempts to succeed. We select tasks where the initial visual state provides sufficient context to elicit correct behavior, leaving language-conditioned specifications to future work. Success rates are measured through human evaluation.

\begin{figure}[t!]
    \centering
    \includegraphics[width=\textwidth]{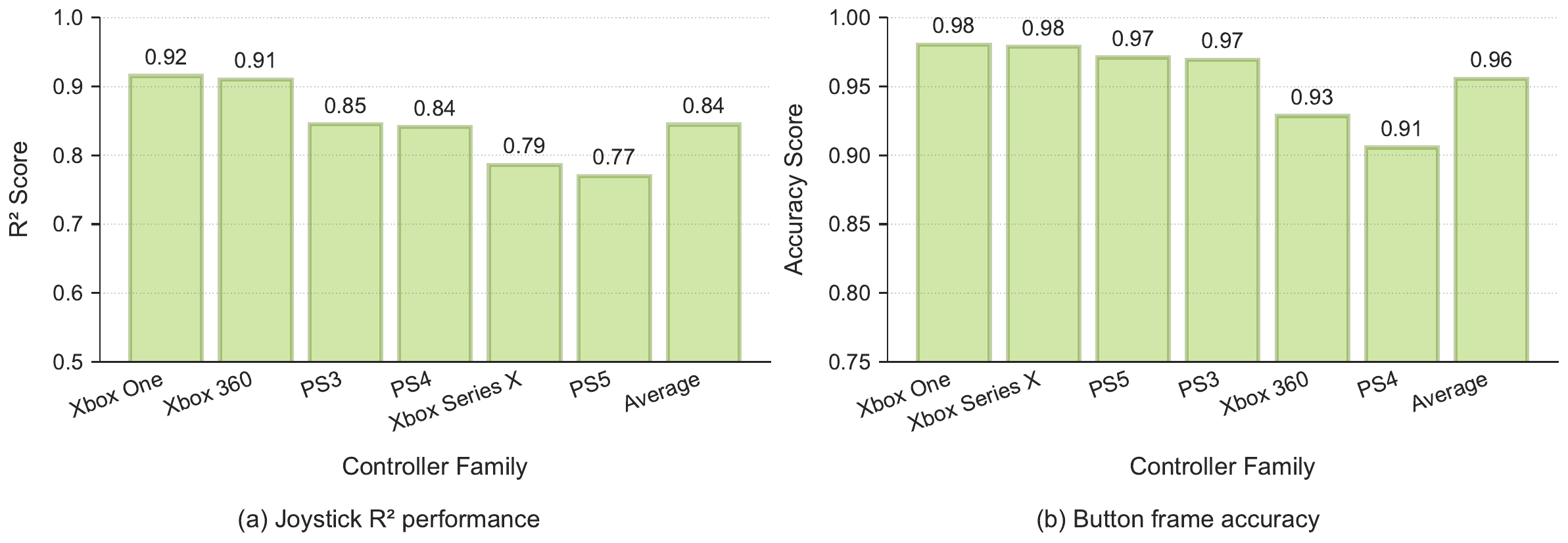}
    \caption{\textbf{Gamepad parsing performance for different controller families.} We verify the correctness of our action extraction pipeline by comparing performance across different controller families against ground-truth data. (a) shows joystick R² correlation scores (averaged for both left and right joysticks) with an overall average of 0.84. (b) shows button frame accuracy with an overall average of 0.96.}
    \label{fig:controller_performance}
\end{figure}

\subsection{\method foundation model}

\textbf{Architecture.}
Building on recent advances in generative modeling and robotics, \method employs flow matching \citep{lipman2022flow} to generate chunks of future actions conditioned on visual observations. The architecture is adapted from GR00T N1 \citep{bjorck2025gr00t} with the language and state encoders removed, and a single action head. RGB inputs at $256 \times 256$ resolution are encoded using a SigLIP 2 vision transformer \citep{tschannen2025siglip}, producing 256 image tokens per frame. Actions are generated with a diffusion transformer (DiT) \citep{peebles2023scalable} that outputs multiple actions per forward pass. Noisy action chunks are first encoded by an MLP into one action token per timestep, then processed through several DiT blocks consisting of alternating self-attention and cross-attention layers. Cross-attention conditions action generation on the encoded frame tokens. The final action tokens are decoded into continuous action vectors using an MLP applied independently across the time dimension. Full mathematical details are provided in Appendix~\ref{app:nitrogen_training}.

\textbf{Design choices.}
Although the model can condition on multiple frames, we find no benefit from using more than one past frame, even with increased temporal gaps. This is likely because the initial state of these action games already provides sufficient context to elicit the appropriate behavior. We instead use a single context frame and generate 16-action chunks, which improves temporal consistency compared to single-action generation. 

\textbf{Training and inference.} 
We train \method using the standard conditional flow-matching objective \citep{lipman2022flow, black2024pi0visionlanguageactionflowmodel}, applied to 16-action chunks, with one $256\times256$ frame of context. Inference follows the corresponding denoising process with $k=16$ steps.

During training, we apply the following image augmentations: random brightness, contrast, saturation and hue, random rotation between $-5$ and $5$ degrees, and random crops. We train all models using AdamW \citep{adam_optimizer, loshchilov2017decoupled} optimizer with a weight decay of $0.001$. We use a warmup-stable-decay (WSD) schedule \citep{wen2024understanding}, which allows us to train for longer without a fixed training budget, with a constant learning rate phase of $0.0001$. Following \cite{peebles2023scalable}, we maintain an exponential moving average (EMA) of model weights during training with a decay of $0.9999$. All our results are obtained with the EMA weights, which we find consistently outperform the non-EMA weights.

\begin{figure}[t!]
    \centering
    
    \includegraphics[width=\textwidth]{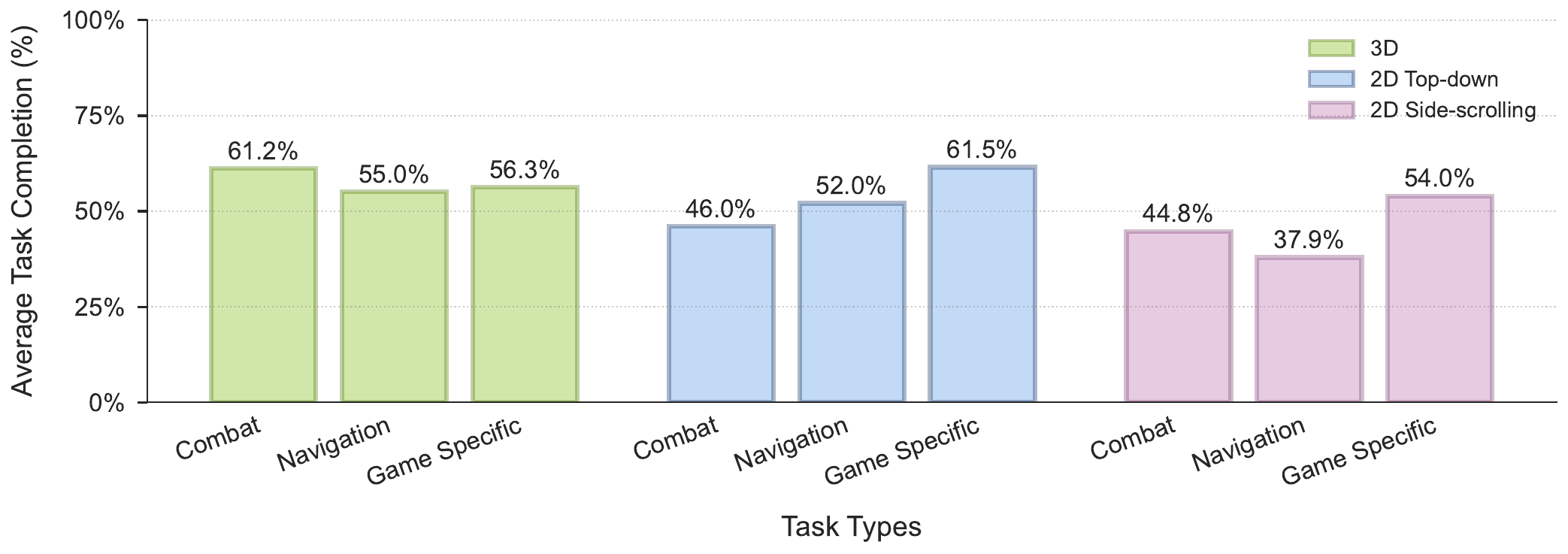}
    \caption{\textbf{\method 500M pre-training results across different games}. We evaluate \method after behavior-cloning pre-training. The model is not fine-tuned for specific games. For each game, we measure the average task completion rate on 3 tasks with 5 rollouts per task. Despite being trained on a very noisy internet dataset, \method is able to perform non-trivial tasks over games with different visual styles (3D, 2D top-down, 2D side-scrolling) and genres (platformer, action-RPG, roguelike, etc.).}
    \label{fig:nitrogen_scores}
\end{figure}

\section{Experiments}

\label{sec:experiments}

\textbf{Performance of the gamepad action extraction model.}
To evaluate our action extraction pipeline, we construct a benchmark dataset by recording gameplay from six video games using OBS\footnote{Open Broadcaster Software: \url{https://obsproject.com/}; Input recording tool: \url{https://github.com/loicmagne/input-rec}}, with randomized opacity, gamepad size, and gamepad type to mimic real-world conditions. We record ground-truth controller inputs at each frame and compare them with the extracted actions. We measure joystick accuracy with the $R^2$ score and button accuracy per frame. As shown in Figure~\ref{fig:controller_performance}, we achieve an average $R^2$ of 0.84 for joystick positions and an average button accuracy of 0.96 across the most popular controller families.

\textbf{\method demonstrates strong capabilities across a wide range of games.}
We train a single model on the entire dataset from Section~\ref{sec:dataset}. Without further fine-tuning, \method achieves non-trivial success rates across many games and tasks. Figure~\ref{fig:nitrogen_scores} summarizes the main results. We observe that \method performs well both on tasks that can be memorized and on tasks that require zero-shot generalization. For example, some games feature fixed layouts that the model may have partially encountered during training, while others employ procedural generation that ensures each playthrough is unique. We do not find significant differences in performance between these two categories, suggesting that \method can both leverage memorization and adapt to unseen scenarios.

This result validates that it is possible to train a robust policy using only noisy internet-scale data. The dataset includes several sources of noise that could hinder training: \textbf{(a) actions are not strictly ground truth}, since input overlay software introduces small delays, and parsing adds further inaccuracies; \textbf{(b) video frames often contain creator-specific artifacts} such as livestream chats, subscribe prompts, or progress trackers; and \textbf{(c) controller configurations vary across players}, differences in sensitivity settings or custom button mappings can change the semantic meaning of the same input. Despite these challenges, Figure~\ref{fig:nitrogen_scores} shows that large-scale pre-training yields a robust multi-game policy.

\begin{figure}[t!]
    \centering
    \includegraphics[width=\textwidth]{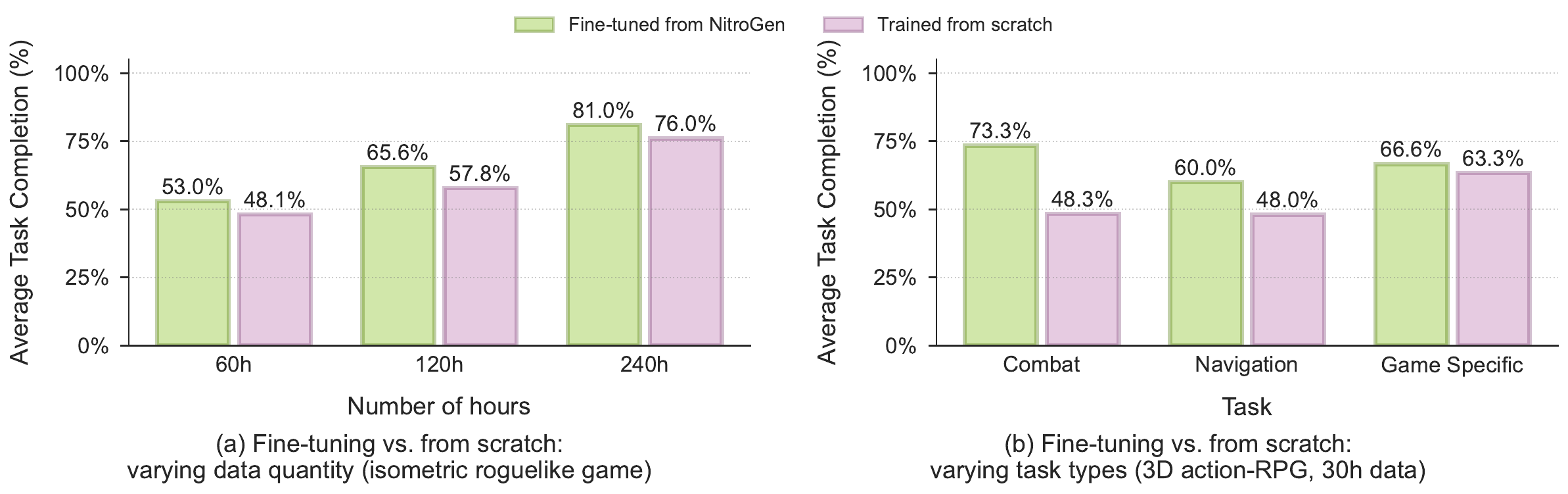}
    \caption{\textbf{Post-training experiments: \method pre-training improves downstream agents in unseen environments.} We pre-train \method on the dataset described in Section \ref{sec:dataset}, holding out one game. We then fine-tune the pre-trained checkpoint on the held-out game and compare the results with a model trained from scratch using the same architecture, data and compute budget. (a) When varying data quantity, task-completion rate scales with dataset size, and fine-tuning achieves on average a 10\% relative improvement in task-completion rate. (b) When varying task type in the low-data regime (30h), fine-tuning achieves up to 52\% relative improvement in task-completion rate.}
    \label{fig:finetuning_results}
\end{figure}

\textbf{\method pre-training improves downstream fine-tuning on unseen environments.}
We evaluate transfer learning by pre-training \method on the full dataset except for a held-out game, then fine-tuning on this game with a limited amount of data. We compare this fine-tuned model with an identical architecture trained from scratch using the same data and compute budget. Results are shown in Figure~\ref{fig:finetuning_results}. We study two representative games with different visual styles and genres: an isometric roguelike and a 3D action-RPG.

The effectiveness of pre-training varies by game type and task category. Across different data quantities, fine-tuning achieves an average relative improvement of 10\% on the isometric roguelike, whereas the 3D action-RPG shows a 25\% average relative improvement. This difference likely stems from better representation of 3D action-RPGs in the training distribution, while the isometric roguelike has gameplay mechanics and visual style that are less common in the training data.

Furthermore, pre-training benefits are not uniform across task types. On the 3D action-RPG, generic tasks such as combat (52\% relative improvement) and navigation (25\% relative improvement) benefit substantially from pre-training, while game-specific tasks show only marginal gains (5\% relative improvement). This suggests that \method effectively learns transferable skills for common gameplay patterns, but game-specific mechanics still require targeted training on the new environment.

\section{Limitations and future work}

\textbf{Design limitations.} \method is limited to being a fast-reacting system-1 sensory model. It cannot plan over long horizons or follow language instructions; the model only reacts to the short context it sees. We develop \method aiming for it to serve as a foundation for future generalist agent development, where post-training for language-following and reinforcement learning can be applied to enhance planning capabilities and improve success rates.

\textbf{Dataset bias.} While diverse, our data collection method still restricts the types of games included in our dataset. The data distribution of the \method dataset is biased toward action games (Figure~\ref{fig:dataset_analysis}), and games that are typically played with a gamepad. Keyboard-only games or those that involve complex manipulation are less represented in the dataset. This bias may limit the agent’s ability to generalize to genres like strategy or simulation games that rely more on planning and keyboard input.

\section{Related works}

\textbf{Gaming agents.} Video games have long been testbeds for AI, with approaches generally following three directions. Reinforcement learning achieved landmark successes from Atari with DQN \citep{mnih2013playing, mnih2015human} to AlphaGo \citep{silver2016mastering}, AlphaStar \citep{vinyals2019grandmaster}, and OpenAI Five \citep{berner2019dota}, but these rely on engineered rewards, hand-crafted features, and specialized simulators. More recent vision-based methods like Dreamer 3 \citep{hafner2023mastering} still require dedicated simulators and environment-specific training. A second line leverages large language models for high-level reasoning with structured APIs, as in Voyager \citep{wang2023voyager} and Cradle \citep{tan2024cradle}, but these depend on hand-crafted interfaces. A third category learns directly from pixels or states via behavior cloning, including MineRL \citep{guss2019minerl}, VPT \citep{baker2022video}, SIMA \citep{sima2024}, GATO \citep{reed2022generalistagent}, Dreamer 4 \cite{hafner2025trainingagentsinsidescalable}, Lumine \cite{tan2025lumineopenrecipebuilding}, and Farhang et al.~\cite{farhang2024humanlike}, but they all rely on datasets bootstrapped from human demonstrations or RL-generated data. \method advances this third direction by scaling behavior cloning to internet-scale, enabling training across hundreds of games without costly collection. Game-TARS \cite{wang2025gametarspretrainedfoundationmodels} is a concurrent work that also train a multi-game agent. They combine contractor data and multi-modal reasoning data to train on a total of 20,000 hours.

\textbf{Embodied foundation models.} Foundation models for embodied AI generally adopt either hierarchical reasoning or end-to-end learning. Hierarchical methods pair pre-trained LLMs or VLMs with low-level policies \citep{ahn2022can, driess2023palme, huang2022inner, liang2023code, singh2023progprompt} treating the foundation models as black-boxes. Vision-Language-Action (VLA) models \citep{bjorck2025gr00t, kim2024openvla, black2024pi0, brohan2022rt, cheang2024gr, wen2025tinyvla, team2024octo} instead train policies end-to-end on embodied data, though generalizing across tasks and embodiments remains challenging. \method differs by discarding language conditioning and focusing purely on scalable vision-action mapping using diverse gameplay data.

\textbf{Large-scale action datasets.} Progress in vision and NLP has been driven by large labeled datasets, but embodied AI lags behind due to the difficulty of collecting action-labeled data and defining standardized action spaces. Gaming datasets like MineRL \citep{guss2019minerl} provide limited coverage, while MineDojo \citep{fan2022minedojo} scales video data without action labels. VPT \citep{baker2022video} annotates 70,000 hours via inverse dynamics but is limited to Minecraft. Other work seeks to infer latent actions from videos \citep{edwards2019imitating, ye2024latent, bruce2024genie, parker2024genie}, though scalability is unclear. In robotics, teleoperation has produced datasets such as Roboturk \citep{mandlekar2018roboturk, mandlekar2019scaling, mandlekar2020human}, ALOHA \citep{aldaco2024aloha}, TeleMoMa \citep{dass2024telemoma}, Open X-Embodiment \citep{o2024open}, and AgiBot World \citep{bu2025agibot}, but these are costly, limited in scale, and lack organic diversity. \method introduces a scalable alternative by leveraging input overlay software, which naturally provides action labels in publicly available gameplay videos.
\section{Conclusion}

In this work, we introduce \method, an approach to scale up foundation pre-training for video game agents and demonstrate how internet pre-training can yield a generalist policy. We leverage a new source of publicly available data to build an internet-scale video-action dataset, and empirically demonstrate its effectiveness by successfully training a multi-game policy. \method shows positive signs of generalization in fine-tuning experiments. By lowering the barrier to train agents on new environments, \method serves as a starting point to develop more powerful and general-purpose agents.

\bibliographystyle{plainnat}  
\bibliography{paper}  

\appendix

\clearpage

\section{NitroGen model details}

\label{app:nitrogen_training}

\subsection{Training objective}

Given a ground-truth action chunk $a \in \mathbb{R}^{16 \times 24}$, an observation $o \in \mathbb{R}^{256 \times 256}$, a flow-matching timestep $t \in [0,1]$, and Gaussian noise $\epsilon \sim \mathcal{N}(0, \mathcal{I})$, we construct the noisy action as
\[
a_t = (1 - t) \cdot \epsilon + t \cdot a
\]
and define the conditional velocity field as
\[
u^{\text{cond}}(x, t, a, \epsilon, o) = a - \epsilon.
\]
The model is trained to predict the velocity field by minimizing the conditional flow-matching loss:
\begin{equation}
    \mathcal{L}^{CFM}(\theta, \phi) = \mathbb{E}_{t, a, \epsilon} \left[ \left\lVert \pi_\theta(a_t, \psi_\phi(o), t) - (a - \epsilon) \right\rVert^2 \right],
    \label{eq:flow_matching_loss}
\end{equation}
where $\pi_\theta$ is the DiT and $\psi_\phi$ is the image encoder. Following \cite{bjorck2025gr00t, black2024pi0visionlanguageactionflowmodel}, we sample $t$ from a shifted beta distribution that prioritizes small timesteps.

\subsection{Inference}
At inference time, we initialize $a_0 \sim \mathcal{N}(0, \mathcal{I})$ and iteratively denoise for $k$ steps using Euler integration:
\begin{equation}
    a_{t+1/k} = a_t + \frac{1}{k} \pi_\theta(a_t, \psi_\phi(o), t).
    \label{eq:denoising}
\end{equation}
We use $k=16$ denoising steps, as additional steps yield no measurable improvement.

\section{Evaluation}

\subsection{Synchronous inference}
\label{app:inference}

As described in Section \ref{sec:gym_api}, we use a Gymnasium API that freezes the game while the model predicts the next action. Frequent pausing and resuming during gameplay could potentially affect the game's physics engine in unknown ways. To rule out this possibility, we record videos and actions (ground truth) of humans playing several games for five minutes each, focusing on parts of the game that are expected to be deterministic (e.g., no enemy behavior randomness). We then replay the same actions from the same initial position: (a) in real time without pausing, and (b) while pausing and resuming the game with random pause durations at high frequency. We find that replayed sequences begin visually diverging after one minute for games with continuous actions and after about three minutes for games with only discrete actions. This result is the same for both (a) and (b), confirming the correctness of our approach: the divergence is simply due to error accumulation.

\end{document}